\documentclass[10pt,twocolumn,letterpaper]{article}

\usepackage[pagenumbers]{cvpr} 

\newcommand\extrafootertext[1]{%
    \bgroup
    \renewcommand\thefootnote{\fnsymbol{footnote}}%
    \renewcommand\thempfootnote{\fnsymbol{mpfootnote}}%
    \footnotetext[0]{#1}%
    \egroup
}

\usepackage{graphicx}
\usepackage{amssymb}
\usepackage{booktabs}
\usepackage{xcolor}         
\usepackage{multirow}
\usepackage{color, colortbl}

\definecolor{Gray}{gray}{0.90}
\usepackage{times}
\usepackage{epsfig}
\usepackage{array}
\renewcommand\arraystretch{1.0}
\usepackage{dblfloatfix}
\usepackage{amsmath}
\usepackage{bm}
\usepackage{breqn}
\usepackage{wrapfig}
\usepackage{adjustbox}
\usepackage[rightcaption]{sidecap}

\usepackage[pagebackref=true,breaklinks=true,colorlinks,bookmarks=false,citecolor=blue,linkcolor=blue,urlcolor=blue]{hyperref}

\usepackage{amssymb}
\usepackage{pifont}
\newcommand{\xmark}{\ding{55}}%
\newcommand{\cmark}{\ding{51}}%

\usepackage[capitalize]{cleveref}
\crefname{section}{Sec.}{Secs.}
\Crefname{section}{Section}{Sections}
\Crefname{table}{Table}{Tables}
\crefname{table}{Tab.}{Tabs.}

\newcommand{\txt}[1]{{\texttt{#1}}}
\usepackage[misc]{ifsym}
\newcommand{\corrAuthor}{$^{\textrm{\Letter}}$}
\usepackage[hang,flushmargin]{footmisc}

\def\cvprPaperID{} 
\def\confName{CVPR}
\def\confYear{2023}

\begin{document}

\title{SwiftFormer: Efficient Additive Attention for Transformer-based \\ Real-time Mobile Vision Applications}

\author{%
  Abdelrahman Shaker$^{1}$\corrAuthor \quad 
  Muhammad Maaz$^{1}$ \quad 
  Hanoona Rasheed$^{1}$ \quad
  Salman Khan$^{1}$ \\
  Ming-Hsuan Yang$^{2,3,4}$ \quad
  Fahad Shahbaz Khan$^{1,5}$
  \vspace{0.5em} \\
  $^{1}$Mohamed bin Zayed University of AI \quad 
  $^{2}$University of California, Merced \\
  $^{3}$Yonsei University \quad
  $^{4}$Google Research \quad
  $^{5}$Link\"{o}ping University 
  }
\maketitle
\extrafootertext{
\textsuperscript{\corrAuthor}\txt{abdelrahman.youssief@mbzuai.ac.ae}}
\begin{abstract}
Self-attention has become a defacto choice for capturing global context in various vision applications. However, its quadratic computational complexity with respect to image resolution limits its use in real-time applications, especially for deployment on resource-constrained mobile devices. Although hybrid approaches have been proposed to combine the advantages of convolutions and self-attention for a better speed-accuracy trade-off, the expensive matrix multiplication operations in self-attention remain a bottleneck. In this work, we introduce a novel efficient additive attention mechanism that effectively replaces the quadratic matrix multiplication operations with linear element-wise multiplications. Our design shows that the key-value interaction can be replaced with a linear layer without sacrificing any accuracy. Unlike previous state-of-the-art methods, our efficient formulation of self-attention enables its usage at all stages of the network. Using our proposed efficient additive attention, we build a series of models called ``SwiftFormer" which achieves state-of-the-art performance in terms of both accuracy and mobile inference speed. Our small variant achieves 78.5\% top-1 ImageNet-1K accuracy with only 0.8~ms latency on iPhone 14, which is more accurate and 2$\times$ faster compared to MobileViT-v2. Code: \url{https://tinyurl.com/5ft8v46w}
\end{abstract}

\section{Introduction}

\noindent In recent years, transformer models have shown remarkable success in various vision applications such as classification~\cite{ViTs, deit, Swin, Swin_v2, MViT}, detection~\cite{DETR, zhu2020deformable, meng2021conditional, zhang2022dino, Maaz2022Multimodal}, and segmentation~\cite{MaskFormer, UNETR++}. However, deploying these models on resource-constrained mobile devices for real-time applications remains challenging due to their inherently complex nature~\cite{EfficientFormer, EdgeNeXt}. Specifically, vision transformers (ViTs) rely on global self-attention, which has a quadratic complexity with respect to the input image resolution, making it impractical for deployment on low-powered mobile devices~\cite{MobileViT2}. As a result, convolutional neural networks (CNNs) are still the preferred choice for real-time deployment on mobile devices, primarily because the convolution operation is computationally efficient~\cite{MobileNetV2, MobileNetV3}. However, a major limitation of CNNs is their reliance on local connections and stationary weights, which can limit their ability to adapt to variable input resolutions and capture long-range dependencies in the data. Therefore, developing more efficient and flexible models that combine the strengths of both CNNs and transformers is critical, particularly for mobile devices with limited computational resources.

\begin{figure}[t]
  \centering
    \includegraphics[width=1.0\linewidth]{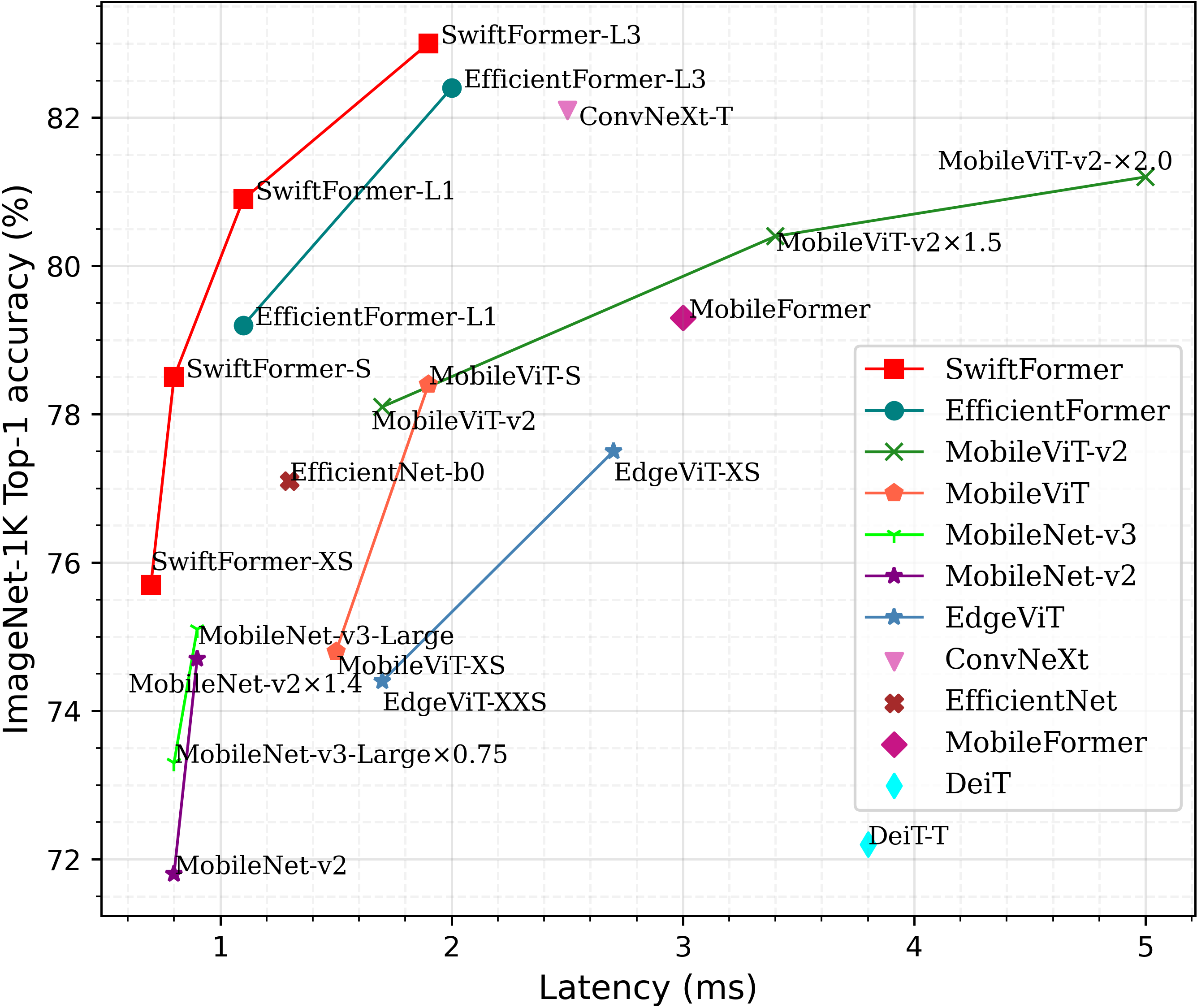}
    \caption{\textbf{Latency vs Accuracy Comparison}. 
    Compared to the recent EfficientFormer-L1~\cite{EfficientFormer}, our SwiftFormer-L1 achieves an absolute gain of 1.7\% in terms of top-1 accuracy with the same latency and without requiring any neural architecture search. 
    }
    \label{fig:introduction}
    \vspace{-0.5cm}
\end{figure}

To achieve this goal, several hybrid approaches have been proposed that use lightweight CNN modules in the high-resolution early stages and self-attention in the low-resolution later stages~\cite{zhang2022parcnet, EdgeNeXt, EfficientFormer}. This approach effectively increases the receptive field of the network and strives to achieve a trade-off between speed and accuracy. Furthermore, different efficient variants of computing self-attention have been proposed to reduce the model complexity. These include computing attention across feature dimensions to implicitly model the global context~\cite{EdgeNeXt}, computing attention within local windows~\cite{Swin}, pooling spatial features before applying self-attention~\cite{MViT}, and sparsely attending to a fixed number of tokens~\cite{EdgeViT}, to name a few.

Although these approaches effectively reduce network complexity, they still involve inefficient matrix multiplication operations that significantly impact latency on mobile devices. To address this issue, Mehta et al.~\cite{MobileViT2} propose a separable self-attention mechanism that replaces matrix multiplication operations to element-wise multiplications. This is achieved by projecting queries to context scores, followed by element-wise multiplication with keys to calculate context vectors for encoding global context.

In this work, we propose efficient additive attention, which eliminates the need for expensive matrix multiplication operations in computing self-attention.~Additionally, we propose to compute the global context using only the query-key interactions followed by a linear transformation, without requiring explicit key-value interactions. 
This significantly reduces the computational complexity and enables us to use the proposed attention block in all stages of the network. Our contributions are as follows:
\begin{itemize}\setlength{\itemsep}{0em}
\item We introduce \emph{efficient additive attention}, a new approach for computing self-attention in vision backbones that eliminates the need for expensive matrix multiplication operations, significantly reducing the computational complexity of the model.

\item Unlike previous methods, our proposed efficient attention design can be used at all stages of the network, enabling more effective contextual information capture and achieving superior speed-accuracy trade-off.

\item We build a series of efficient generic classification models called ``SwiftFormer", which utilize our proposed \emph{efficient additive attention}. Our \textit{small} model achieves 78.5\% top-1 ImageNet-1K~\cite{deng2009imagenet} accuracy while running at only 0.8~ms latency on iPhone 14. Moreover, our large model achieves 83.0\% accuracy with a latency of only 1.9~ms. Our model achieves state-of-the-art performance, outperforming recent MobileViT-v2~\cite{MobileViT2} and EfficientFormer~\cite{EfficientFormer} by obtaining a better trade-off between accuracy and latency~(see Fig.~\ref{fig:introduction}).

\end{itemize}

\section{Related Work}
\noindent
\textbf{Efficient CNNs}: Designing efficient CNNs for mobile vision applications has received much attention in recent years. MobileNet architectures ~\cite{MobileNet, MobileNetV2, MobileNetV3} propose depth-wise separable convolutions as well as efficient inverted residual blocks for improved performance on various vision tasks. Other methods aim to improve the efficiency by leveraging depth-wise dilated convolutions ~\cite{espnetv2}, channel shuffling and pointwise group convolutions ~\cite{ShuffleNet, ShuffleNetV2}, network pruning ~\cite{DCompress, Pruning}, low bit-width ~\cite{BitMix, Jacob_2018_CVPR}, and neural architecture search ~\cite{mnasnet, MobileNetV3}. CNN-based methods are well-performing, efficient, and fast to train and run on edge devices, resulting in widespread usage in the industry. However, they are spatially local and lack global interaction between the features, which deeply affects their performance.
\noindent
\textbf{Efficient Transformers}: ViTs~\cite{ViTs} have been widely used in numerous vision tasks, and significant advances have been made in terms of data efficiency~\cite{liu2021efficient, deit}, transformer architecture \cite{MobileViT, EfficientFormer, MobileFormer, EdgeFormer}, and token mechanisms \cite{tolstikhin2021mlp, MetaFormer}. 
Reducing the number of visual tokens is a major modification in the transformer architecture for efficient deployment. 
Instead of using a fixed feature representation through the whole architecture, some methods employ a hierarchical design where the resolution is gradually decreased through the stages, including down-sampling techniques~\cite{MViT, UNETR++, TokenLearner, PiT} and pyramidal structures~\cite{PyramidViT, PyramidViT2}. 
Recently, a few methods~\cite{Token_Sparse, ATS, EdgeViT} propose token sparsification techniques to encode only a subset of the most informative tokens.

Numerous approaches have recently been proposed to reduce the quadratic complexity of self-attention, the computational bottleneck in transformer-based architectures, 
by computing its approximated variants \cite{EdgeViT, EdgeNeXt, MobileViT2, LinFormer, ReFormer, Twins, MaxViT}.
EdgeViT~\cite{EdgeViT} uses a global sparse attention module attending only to a few tokens to improve the efficiency, while~\cite{pvt} down-samples the key and value vectors that lead to a better efficiency-accuracy trade-off.
EdgeNeXt~\cite{EdgeNeXt} adopts transposed self-attention operation to compute the attention maps across the channel dimension instead of the spatial dimension, followed by token mixing, to have a linear complexity with respect to the number of tokens. Reformer~\cite{ReFormer} replaces the dot-product attention with a locality-sensitive hashing to group the tokens and reduced the complexity from $O(n^2)$ to $O$($n$ $\log n$). 
However, this design is only efficient on longer sequences, which is typically not the case for ViTs.
LinFormer~\cite{LinFormer} is a low-rank matrix factorization method that approximates the self-attention matrix with a low-rank matrix, reducing the complexity from $O(n^2)$ to $O$($n$). 
Although matrix factorization methods theoretically reduce the complexity of self-attention, they use expensive projections for computing attention, which may not reflect the reduction in FLOPs and parameters into actual speed on mobile platforms. 

Although these methods show promise and have reduced the complexity of self-attention
theoretically, they are inadequate for reducing the inference speed for mobile deployment. 
Since the complexity of the multi-headed self-attention (MHSA) is higher in the earlier stage compared to the last stages,
EfficientFormer~\cite{EfficientFormer} incorporates MHSA in the last stage only to learn contextual information from the high-level features without increasing the inference speed significantly. 
Recently,
MobileViT-v2~\cite{MobileViT2} proposes separable self-attention that uses element-wise operations instead of the dot-product to compute the attention maps with linear complexity. 
Different from the existing approaches, we propose a consistent hybrid design with an \textit{efficient additive attention} mechanism to model the contextual information with linear complexity. 
Instead of capturing the pair-wise interactions between keys, queries, and values using the dot-product, we use element-wise operations with learnable attention weights to model the interactions between query and keys only, leading to better inference speed.

\section{Method}

\noindent\textbf{Motivation}: 
To motivate our method, we first distinguish three desirable characteristics to be considered when designing an efficient yet accurate approach for resource constraint mobile devices. 

\noindent\textbf{\textit{Efficient Global Context Modeling}}: 
As discussed earlier, most existing approaches either employ the standard MHSA or an approximated variant to learn the global context. However, they struggle to operate as fast as MobileNets on resource-constrained devices. This is likely due to the computation-intensive multiplicative operations during attention computation or reliance on advanced reshaping and indexing operations in these approaches. For instance, the recent MobileViT-v2~\cite{MobileViT2} is 2$\times$ slower than MobileNet-v2~\cite{MobileNetV2}. Instead of using matrix multiplications, we argue that encoding the global context using an efficient additive attention design can reduce the operations with respect to the number of tokens. This is expected to help operate at comparable speed and model size, while achieving superior accuracy compared to MobileNets.

\noindent\textbf{\textit{Rethinking key-value interactions}}: 
Other than multiplicative operations during attention computation, additive attention has been recently explored in the NLP domain \cite{fastformer}. However, in the standard form, it performs three-step processing to model query, key, and value interactions. Each step feeds into the subsequent one, thereby requiring sequential processing. Here, we rethink the additive attention for vision tasks by alleviating the need to compute explicit interactions between key-value. We argue that eliminating key-value interactions and replacing them with a simple linear transformation empirically encodes better contextual representation. Our design encodes only global queries and key interactions to learn the global contextual information, followed by a linear transformation to compute global context-aware attention weights.

\noindent\textbf{\textit{Consistent Hybrid Design}}: Most existing works employ MHSA or the approximated variant in the last stages, while avoiding its usage in the earlier stages. This is because the computational complexity of MHSA grows quadratically with the length of the tokens, making it impractical to incorporate during the initial stages. This constraint adds to the design complexity and requires a careful selection of stages, where MHSA can be applied. In contrast, our proposed SwiftFormer module has linear complexity with respect to the token length and can be incorporated in all stages to learn consistent global context at each scale. This consistency improves the model performance and makes it more generalizable and scalable for high-resolution images.

\subsection{Overview of Attention Modules}
\begin{figure*}[t]
  \centering
    \includegraphics[width=1.0\linewidth]{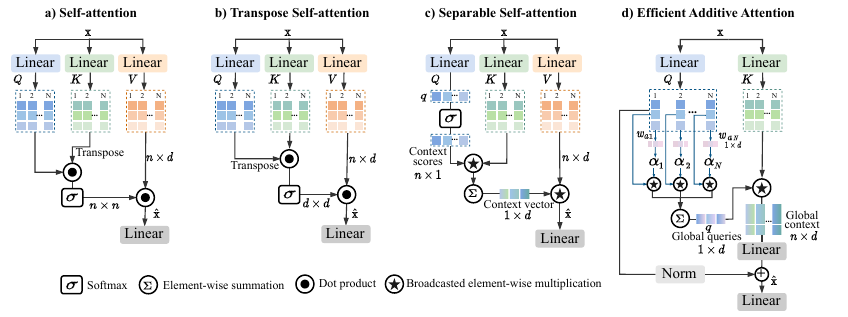}
    \caption{\textbf{Comparison with different self-attention modules}. \textbf{(a)} is a typical self-attention used in ViTs~\cite{ViTs}. \textbf{(b)} is the transpose self-attention used in EdgeNeXt ~\cite{EdgeNeXt}, where the self-attention operation is applied across channel feature dimensions ({d}$\times${d}) instead of the spatial dimension ({n}$\times${n}).  \textbf{(c)} is the separable self-attention of MobileViT-v2~\cite{MobileViT2}, it uses element-wise operations to compute the context vector from the interactions of \textbf{Q} and \textbf{K} matrices. Then, the context vector is multiplied by \textbf{V} matrix to produce the final output. \textbf{(d)} Our proposed efficient additive self-attention. Here, the query matrix is multiplied by learnable weights and pooled to produce global queries. Then, the matrix \textbf{K} is element-wise multiplied by the broadcasted global queries, resulting the global context representation.}
    \label{fig:Attention_Comparison}
\end{figure*}

\noindent
Vision transformer models are built upon the self-attention (see Fig.~\ref{fig:Attention_Comparison}~(a)), which can effectively model the interactions between the input tokens. Specifically, the self-attention has $\mathbf{x}$ as an input, where $\mathbf{x} \in \mathbb{R}^{n \times d}$, comprising $n$ tokens with $d$-dimensional embedding vector. The input $\mathbf{x}$ is projected to query ($\mathbf{Q}$), key ($\mathbf{K}$), and value ($\mathbf{V}$) using three matrices, $\mathbf{W_Q}$, $\mathbf{W_K}$, and $\mathbf{W_V}$. Each self-attention layer
comprises $h$ heads, which allows the model to attend to different views of the input. The 
self-attention can be described as:
 \begin{equation}   \hat{\mathbf{x}}=\texttt{Softmax}\Big(\frac{\mathbf{Q} \cdot \mathbf{K}^\top} {\sqrt{d}}\Big)\cdot \mathbf{V}.
    \label{eq:SA}
\end{equation}
The attention scores between each pair of tokens in $\mathbf{Q}$ and $\mathbf{K}$ are computed using the dot-product operation. Next, these scores are normalized followed by Softmax to weigh the interactions between the tokens. Finally, the weighted interactions are multiplied by $\mathbf{V}$ using the dot-product operation to produce the final weighted output. Overall, the complexity of the self-attention is $O(n^2 \cdot d)$, where $n$ is the number of tokens and $d$ is the hidden dimension. The computational and memory demands of $\mathbf{Q} \cdot\mathbf{K}^\top$ increase quadratically as the number of tokens grows, leading to slow inference speed and high memory usage, making it impractical to run in real-time for long sequences.

To alleviate this issue, ~\cite{EdgeNeXt} proposes the transpose self-attention (see Fig.~\ref{fig:Attention_Comparison}~(b)) to reduce the complexity from quadratic to linear with respect to the number of tokens. Here, the dot-product operation is applied across the channel dimension instead of the spatial dimension. This allows the model to learn feature maps with implicit contextual representation. 
The attention can be described as:
\begin{equation} \hat{\mathbf{x}}=\mathbf{V}\cdot\texttt{Softmax}\Big(\frac{\mathbf{Q}^{\top}\cdot\mathbf{K}}{\sqrt{d}}\Big).
    \label{eq:CA}
\end{equation}
The transpose self-attention has a computational complexity of $O(n \cdot d^2)$. While this complexity scales linearly with the number of tokens $n$, it remains quadratic with respect to the feature dimension $d$. Further, the dot-product operation is still utilized between the query and key matrices.

The separable self-attention mechanism (see Fig.~\ref{fig:Attention_Comparison}~(c)) aims to address the bottleneck of the standard self-attention. Here, the interactions between the queries ($\mathbf{Q}$), keys ($\mathbf{K})$, and values ($\mathbf{V}$) are encoded using element-wise operations. First, the query matrix $\mathbf{Q}$ is projected to produce a vector $\mathbf{q}$ of dimensions $n\times1$, and then fed into Softmax to generate the context scores, which captures the importance of each query element. Then, the context scores are multiplied by the key matrix $\mathbf{K}$ and pooled to compute a context vector, which encodes the contextual information. Finally, the context vector is multiplied element-wise with the value matrix $\mathbf{V}$ to propagate the contextual information and produce the final output $\hat{\mathbf{x}}$. It can be summarized as:
\begin{equation}
  \hat{\mathbf{x}}= \mathbf{V} * \sum  \mathbf{K} * \texttt{Softmax}(\mathbf{q}). 
\end{equation}
Here, $*$ denotes the element-wise multiplication operation.

\subsection{Efficient Additive Attention}
\label{sec:EAA}

\noindent The typical additive attention mechanism in NLP captures the global context by utilizing pairwise interactions between the tokens via element-wise multiplications instead of using dot-product operation. It encodes the relevance scores for the contextual information of the input sequence based on the interactions of the three attention components ($\mathbf{Q}$, $\mathbf{K}$, $\mathbf{V}$). In contrast, we show that key-value interactions can be removed without sacrificing the performance and only focusing on effectively encoding query-key interactions by incorporating a linear projection layer is sufficient to learn the relationship between the tokens (see Fig.~\ref{fig:Attention_Comparison}~(d)). This approach, named efficient additive attention, has a faster inference speed and produces more robust contextual representations as demonstrated by our performance on image classification, object detection, and segmentation tasks (Sec. \ref{sec:results}). Specifically, the input embedding matrix $\mathbf{x}$ is transformed into query ($\mathbf{Q}$) and key ($\mathbf{K}$) using two matrices $\mathbf{W_q}$, $\mathbf{W_k}$, where $\mathbf{Q}$, $\mathbf{K}$ $\in \mathbb{R}^{n \times d}$, $\mathbf{W_q}$, $\mathbf{W_k}$ $\in \mathbb{R}^{d \times d}$, $n$ is the token length and $d$ is the dimensions of the embedding vector. Next, the query matrix $\mathbf{Q}$ is multiplied by learnable parameter vector $\mathbf{w}_a$ $\in \mathbb{R}^{d}$ to learn the attention weights of the query, producing global attention query vector $\alpha$ $\in \mathbb{R}^{n}$ as:

\begin{equation}
\alpha=\mathbf{Q}\cdot\mathbf{w}_a/\sqrt{d}
\end{equation}

Then, the query matrix is pooled based on the learned attention weights, resulting in a single global query vector $\mathbf{q}$ $\in \mathbb{R}^{d}$ as follows:
 \begin{equation}
    \mathbf{q}=\sum_{i=1}^{n} \alpha_{i} * \mathbf{Q_i}.
\label{eq:Additive_Attention_Q}
\end{equation}
Next, the interactions between the global query vector $\mathbf{q}$ $\in \mathbb{R}^{d}$~and key matrix $\mathbf{K}$ $\in \mathbb{R}^{n\times d}$ are encoded using the element-wise product to form global context ($\mathbb{R}^{n\times d}$).
This matrix shares a similarity with the attention matrix in MHSA and captures information from every token and has the flexibility to learn the correlation in the input sequence. However, it is comparatively inexpensive to compute compared to MHSA and has linear complexity with the token length. Inspired by the transformer architecture, we employ a linear transformation layer to query-key interactions to learn the hidden representation of the tokens.
The output of the efficient additive attention $\hat{\mathbf{x}}$ can be described as:
\begin{equation}
  \hat{\mathbf{x}}= \hat{\mathbf{Q}} + \mathbf{T}\Big(\mathbf{K} *\mathbf{q}\Big). 
  \label{sec:reduced_additive_attention}
\end{equation}
where $\hat{\mathbf{Q}}$ denotes to the normalized query matrix, $\mathbf{T}$ denotes to the linear transformation.

\begin{figure}[t]
  \centering
    \includegraphics[width=1.0\linewidth]{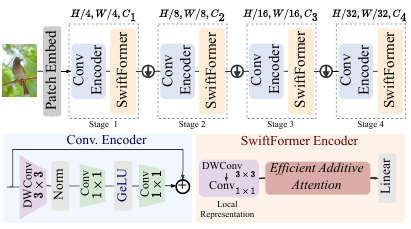}
    \caption{\textbf{Top Row:} Overview of our proposed architecture. The input image is fed into the patch embedding layer, followed by hierarchical stages at four different scales $\{\frac{1}{4},\frac{1}{8},\frac{1}{16},\frac{1}{32}\}$. Each stage is consistent and compose of Conv. Encoder blocks followed by SwiftFormer Encoder. Between two consecutive stages, we incorporate downsampling layer to reduce the spatial size by a factor of two and increase the feature dimensions. \textbf{Bottom Row:} We show the design of the Conv. Eencoder (left) and the SwiftFormer Encoder (right). The Conv. Encoder is designed to learn effective local representations and consists of 3$\times$3 depth-wise convolutions followed by two point-wise convolutions for channel mixing. The SwiftFormer Encoder aims to learn enriched local-global representations. It begins with local convolutional layers to extract local features, followed by the efficient additive attention module (see Fig.~\ref{fig:Attention_Comparison}~(d)) and linear layers.}
    \label{fig:Architecture}
\end{figure}

\subsection{SwiftFormer Architecture} 
\noindent  Our SwiftFormer is based on the recently introduced EfficientFormer~\cite{EfficientFormer}. The main idea of EfficientFormer is to introduce 4D MetaBlocks based on PoolFormer~\cite{PoolFormer} to learn local representations efficiently, while using 3D MetaBlocks based on self-attention to encode global context. However, the performance of EfficientFormer is limited by two design choices. Firstly, it uses ineffective token mixing, and secondly, it only employs 3D MetaBlocks in the last stage due to quadratic complexity of MHSA. This likely leads to inconsistent and insufficient contextual representation. To address these limitations, our SwiftFormer improves the token mixing by using a simple yet effective Conv.~Encoder. Further, we introduce efficient additive attention module that can be incorporated in all stages (Sec. \ref{sec:EAA}). This leads to more consistent learning of local-global representations. It is worth mentioning that EfficientFormer employs a latency-driven slimming method to obtain optimal configurations for its model variants, which leads to maximizing the speed. In contrast, our SwiftFormer models are built without using any neural architecture search.

Fig.~\ref{fig:Architecture} shows an overview of our proposed architecture. The main components are: \textbf{(i)} Effective Conv. Encoder, and \textbf{(ii)} SwiftFormer Encoder. In contrast to other hybrid designs, the proposed architecture is consistent and has Conv.~Encoders followed by SwiftFormer Encoder in all stages. Our architecture extracts hierarchical features at four different scales across four stages. At the beginning of the network, the input image of size $H{\times}W{\times}3$ is fed through \texttt{Patch Embedding} layer, implemented with two $3{\times}3$ convolutions with a stride of 2, resulting $\frac{H}{4}{\times}\frac{W}{4}{\times}C_{1}$ feature maps. Then, the output feature maps are fed into the first stage, which begins with Conv.~Encoder to extract spatial features, followed by SwiftFormer to learn the local-global information. Between two consecutive stages, there is a downsampling layer to increase the channel dimension and reduce the token length. Next, the resulting feature maps 
are subsequently fed into the second, third, and fourth stages of the architecture, producing $\frac{H}{8}{\times}\frac{W}{8}{\times}C_{2}$, $\frac{H}{16}{\times}\frac{W}{16}{\times}C_{3}$, and $\frac{H}{32}{\times}\frac{W}{32}{\times} C_{4}$ dimensional feature maps, respectively. Hence, each stage learns local-global features at different scales of the input image, which allows the network to have enriched representation.

\noindent\textbf{\textit{Effective Conv.~Encoder}}: The baseline EfficientFormer \cite{EfficientFormer} employs 3 $\times$ 3 average pooling layers as a local token mixer, similar to PoolFormer~\cite{PoolFormer}. Although PoolFormer layers are known for their fast inference speed, replacing them with depth-wise convolutions does not increase the latency. Further, it improves the performance without increasing the parameters and latency. Specifically, the features maps $\mathcal{X}_i$ are fed into 
  $3\times3$ depth-wise convolution ($\texttt{DWConv}$) followed by Batch Normalization ($\texttt{BN}$). Then, the resulting features are fed into two point-wise convolutions ($\texttt{Conv}_{1}$) alongside $\texttt{GeLU}$ activation. 
  Finally, we incorporate a skip connection to enable information to flow across the network. The Conv. Encoder is defined as:
\begin{equation}
    \mathcal{\hat{X}}_i = \texttt{Conv}_{1}(\texttt{Conv}_{1,G}(\texttt{DWConv}_{BN}(\mathcal{X}_i))) + \mathcal{X}_i.
\end{equation}
where $\mathcal{X}_i$ refers to the input features, $\texttt{Conv}_{1,G}$ refers to point-wise convolution followed by $\texttt{GeLU}$, $\texttt{DWConv}_{BN}$ refers to depth-wise convolution followed by Batch Normalization, and $\mathcal{\hat{X}}_i$ refers to the output feature maps. 

\noindent\textbf{\textit{SwiftFormer Encoder}}:
This module is carefully designed to efficiently encode enriched local-global representation in each stage. As shown in Fig.~\ref{fig:Architecture}, the initial block of the SwiftFormer Encoder is composed of $3\times3$ depth-wise convolution followed by point-wise convolution, which enables the module to learn spatial information and encode local representation. Then, the resulting feature maps are fed into the efficient additive attention block, which aims to learn contextual information at each scale of the input size. Finally, the output feature maps are fed into a \texttt{Linear} block, which composes of two $1{\times}1$ point-wise convolution layers, Batch Normalization, and $\texttt{GeLU}$ activation to generate non-linear features. The SwiftFormer Encoder is described as:
\begin{equation}
\begin{gathered}
    \mathcal{\hat{X}}_i = \texttt{Conv}_{1}(\texttt{DWConv}_{BN}(\mathcal{\hat{X}}_i)), \\
    \mathcal{\hat{X}}_i = \texttt{QK}(\mathcal{\hat{X}}_i) + \mathcal{\hat{X}}_i, \\
    \mathcal{\hat{X}}_{i+1} = \texttt{Conv}_{1}(\texttt{Conv}_{BN,1,G}(\mathcal{\hat{X}}_i)) + \mathcal{\hat{X}}_i.
\end{gathered}
\end{equation}
where $\texttt{Conv}_{BN,1,G}$ denotes to Batch Normalization, followed by, $1{\times}1$ $\texttt{Conv}$ layer, followed by $\texttt{GeLU}$, and $\texttt{QK}$ denotes the efficient additive attention (explained in Sec.~\ref{sec:reduced_additive_attention}).

\section{Experiments}
\label{sec:results}
\noindent  We evaluate our SwiftFormer models across four downstream tasks: classification on ImageNet-1K~\cite{deng2009imagenet}, object detection and instance segmentation on MS-COCO 2017~\cite{COCO}, and semantic segmentation on ADE20K~\cite{ADE20K}.

\subsection{Implementation Details}
\noindent
\textbf{ImageNet-1K~\cite{deng2009imagenet}}: All of our models are trained from scratch on ImageNet-1K dataset for 300 epochs with AdamW optimizer~\cite{AdamW} and cosine learning rate scheduler with an initial learning rate of 1e$^{-3}$. We use a linear warm-up for 5 epochs.
We use an image resolution of 224$\times$224 for both training and testing. Following the training recipe of~\cite{EfficientFormer}, we use the same teacher model for distillation~\cite{RegNet}. The experiments are conducted with PyTorch 1.12~\cite{PyTorch} using 8 NVIDIA A100 GPUs. The latency is measured using iPhone 14 (iOS 16), and the throughput is measured using A100 40 GB GPU. For latency measurements, we compile the models using CoreML library~\cite{coreml2021} and perform inference with a batch size of 1. For the throughput on A100, the inference is performed using a batch size of 128.

\noindent
\textbf{MS-COCO 2017~\cite{COCO}}: We use our ImageNet pre-trained models as the backbones in Mask-RCNN framework for object detection and instance segmentation on MS-COCO 2017 dataset. The dataset contains 118K training and 5K validation images. Following~\cite{EfficientFormer}, we finetune our models for 12 epochs with an image size of 1333 $\times$ 800 and batch size of 32 using AdamW optimizer. We use learning rate of 2$e^{-4}$ and report the performance for detection and instance segmentation in terms of mean average precision (mAP).

\noindent
\textbf{ADE20K~\cite{ADE20K}}:
The dataset comprises 20K training and 2K validation images and contains 150 class categories for scene parsing. Similar to~\cite{EfficientFormer} we use our ImageNet pre-trained models to extract image features and semantic FPN~\cite{FPN} as a decoder for segmentation. The model is trained with an image size of 512 $\times$ 512 for 40K iterations with a batch size of 32 using AdamW optimizer. We use poly learning rate scheduling with an initial learning rate of 2e$^{-4}$. We report the semantic segmentation performance in terms of mean intersection over union (mIoU).

\subsection{Baseline Comparison}
\noindent
Table \ref{tab:baseline_comparison} illustrates the impact of integrating our proposed contributions into the baseline EfficientFormer-L1~\cite{EfficientFormer} model in terms of ImageNet-1K top-1 accuracy and inference speed. The first row shows the results of the baseline model, which only includes the self-attention based transformer block in the final stage of the network and achieves a top-1 accuracy of 79.2\% with a latency of 1.1 ms on an iPhone 14 mobile device. The second row replaces the pool mixers in the baseline model with our proposed Conv.~Encoder, resulting in an improvement in performance to 79.9\% while maintaining the same latency. In the third row, we replace the self-attention of the transformer block in the baseline by our proposed efficient additive attention module. Although the performance drops by 0.2\%, the inference speed improves by 0.1 ms, and the model has linear complexity with the number of tokens.
This enables us to integrate the SwiftFormer Encoder that built on the efficient additive attention into all stages and achieve better performance while maintaining the same inference speed as of baseline (first versus last row).

\begin{table}[!ht]
\centering
\resizebox{\columnwidth}{!}{
\begin{tabular}{l *{3}{c}}
  \toprule
 Method & Latency (ms) & Top-1 (\%) \\
 \midrule
 \midrule
   EfficientFormer-L1 (Baseline) & 1.1 & 79.2 \\
    + Replace Pool Mixer by effective Conv. Encoder & 1.1 &  79.9 \\
     + Replace Self-attention by efficient additive attention & 1.0 & 79.7 \\
    \rowcolor{gray!10}+ Incorporate SwiftFormer block across all stages & \textbf{1.1} &  \textbf{80.9} \\
  \bottomrule                             
\end{tabular}
\label{tab:baseline_comparison}
} 
\vspace{0.3em}
\caption{Baseline comparison between our SwiftFormer-L1 and EfficientFormer-L1~\cite{EfficientFormer} on the ImageNet-1K dataset. The latency is measured on iPhone14 Neural Engine.}
\label{tab:baseline_comparison}
\end{table}

\begin{table*}[t]
\label{tab:comparison}
\centering
\small
\scalebox{0.85}{
\begin{tabular}{cccccccc}
\toprule

\rowcolor{gray!10} Model  & Type     & Latency (ms) $\downarrow$ & Throughput (A100) $\uparrow$  & Params(M) $\downarrow$ & GMACs $\downarrow$ & Neural Search  & Top-1(\%) $\uparrow$ \\ 
\hline
\hline
MobileNet-v2$\times 1.0$~\cite{MobileNetV2}              &   ConvNet &   {0.8}    &   9889  &   3.5   &   0.3 &   \xmark  &   71.8    \\

\rowcolor{gray!10} MobileNet-v3-Large$\times 0.75$~\cite{MobileNetV3}       &   ConvNet &   {0.8}    &   10934   &   4.0   &   0.2 &   \cmark  &   73.3     \\

EdgeViT-XXS~\cite{EdgeViT}                              &   Hybrid  &   {1.7}     &   5965    &   4.1    & 0.6 & \xmark  & 74.4        \\

\rowcolor{gray!10} MobileViT-XS~\cite{MobileViT}   & Hybrid   & {1.5}   & 3707   &  2.3    & 0.7 & \xmark  &  74.8       \\

\textbf{SwiftFormer-XS}   & \textbf{Hybrid}    & \textbf{{0.7}}   & \textbf{6034}  & \textbf{3.5}    & \textbf{0.6} & \xmark   & \textbf{75.7}        \\
\hline

\rowcolor{gray!10} MobileNet-v2$\times 1.4$~\cite{MobileNetV2}     & ConvNet    & {0.9}  &  7447  &  6.1    & 0.6   & \xmark & 74.7       \\

MobileNet-v3-Large~\cite{MobileNetV3}    & ConvNet    & {0.9}  &  10351  & 5.4    & 0.3   & \cmark & 75.1      \\

\rowcolor{gray!10} EfficientNet-b0~\cite{EfficientNet}       &   ConvNet &   {1.3}   &  8537   &   5.3   &   0.4 &   \cmark  &   77.1     \\

DeiT-T~\cite{deit}          & Transformer & 3.8 & 5860 & 5.7      & 3.8         & \xmark    &    72.2 \\

\rowcolor{gray!10} EdgeViT-XS~\cite{EdgeViT}   & Hybrid    & {2.7}      & 4812  &  6.7    & 1.1 & \xmark  & 77.5        \\

MobileViT-v2$\times 1.0$~\cite{MobileViT2}     & Hybrid    & {1.7}    & 3201  &  4.9    & 1.8 & \xmark  & 78.1      \\

\rowcolor{gray!10} \textbf{SwiftFormer-S}     & \textbf{Hybrid}     & \textbf{{0.8}}  & \textbf{5051}  &  \textbf{6.1}    & \textbf{1.0} & \xmark  & \textbf{78.5}       \\

\hline

MobileFormer-508M~\cite{MobileFormer}     & Hybrid    & {3.0}  &  4443  & 14.0    & 0.5 & \xmark  & 79.3      \\

\rowcolor{gray!10}PoolFormer-S12~\cite{PoolFormer}     & Pool    & {1.2}  &  3227  & 12.0    & 1.8 & \xmark  & 77.2      \\

EfficientFormer-L1~\cite{EfficientFormer}    & Hybrid    & {1.1}  & 5046  &  12.3    & 1.3 & \cmark  & 79.2      \\

\rowcolor{gray!10}MobileViT-v2$\times 1.5$~\cite{MobileViT2}     & Hybrid    & {3.4}  & 2356 &  10.6    & 4.0 & \xmark  & 80.4      \\

\textbf{SwiftFormer-L1}     & \textbf{Hybrid}    & \textbf{{1.1}}    & \textbf{4469}  & \textbf{12.1}    & \textbf{1.6} & \xmark  & \textbf{80.9}      \\
\hline

\rowcolor{gray!10}ResNet-50~\cite{resnet}    & ConvNet   & 1.9        & 4835 & {25.5}   & 4.1 & \xmark  & 78.5      \\

 PoolFormer-S36~\cite{PoolFormer}     & Pool    & {2.8}  &  1114  & 31.0    & 5.0 & \xmark  & 81.4      \\

\rowcolor{gray!10}ConvNeXt-T~\cite{ConvNeXt}    & ConvNet    & 2.5       & 3235 & {28.6}   & 4.5 & \xmark  & 82.1   \\

 DeiT-S~\cite{deit}          & Transformer & 9.9 & 2990 & 22.5      & 4.5         & \xmark    &    81.8 \\
\rowcolor{gray!10}Swin-T~\cite{ConvNeXt}    & Transformer   & NA      &  2635 & {28.3}   & 4.5 & \xmark  & 81.3   \\

 MobileViT-v2$\times 2.0$~\cite{MobileViT2}     & Hybrid    & {5.0}  & 1906  &  18.5    & 7.5 & \xmark  & 81.2      \\

\rowcolor{gray!10}EfficientFormer-L3~\cite{EfficientFormer}    & Hybrid   & 2.0  & 2691 & {31.3}   & 3.9 & \cmark  & 82.4      \\

\textbf{SwiftFormer-L3}     & \textbf{Hybrid}    & \textbf{{1.9}}  & \textbf{2890} &\textbf{28.5 }     &  \textbf{4.0}   & \xmark  & \textbf{83.0}      \\

\bottomrule
\end{tabular}
}

\vspace{0.1cm}
\caption{\textbf{Comparison of our proposed SwiftFormer with the state-of-the-art counterpart models on ImgeNet-1K.} The latency is measured on iPhone 14 Neural Engine (iOS 16) and the throughput is measured on Nvidia A100 GPU. Our models run faster than MobileNets, Hybrid, and Transformer models, with a better trade-off between accuracy and model complexity. The error for the latency measurement is less than $\pm 0.1$ ms. Our results are shown in bold for all model variants.}
\label{tab:SOTA}
\end{table*}

\subsection{Image Classification}
\noindent Table~\ref{tab:SOTA} presents a comparison of our proposed SwiftFormer models (XS, S, L1, and L3) with previous state-of-the-art ConvNets, transformer-based, and hybrid models. We show that our models set new state-of-the-art results, and outperform the recently introduced EfficientFormer~\cite{EfficientFormer} and MobileViT-v2~\cite{MobileViT2} in all model variants. This comprehensive evaluation shows the advantage of our proposed models in terms of both accuracy and latency on mobile devices.

\noindent
\textbf{Comparison with ConvNets:}
Our SwiftFormer models surpass the widely used lightweight CNNs counterparts significantly in terms of top-1 accuracy, while running faster than the highly optimized MobileNet-v2 and MobileNet-v3 on an iPhone 14 mobile device. Specifically, our SwiftFormer-XS runs 0.1 ms faster than MobileNet-v2$\times$1.0 and MobileNet-v3-Large$\times$0.75 and achieve better top-1 accuracy with a margin of 3.9\% and 2.4\% respectively. Our SwiftFormer-S runs faster than EfficientNet-b0~\cite{EfficientNet} by 1.6$\times$ and achieves 1.4\% higher top-1 accuracy. Further, our SwiftFormer-L3 achieves 4.5\% and 0.9\% gain in top-1 accuracy over ResNet-50 and ConvNeXt-T, respectively, while running at the same latency as ResNet-50 and 1.3$\times$ faster than ConvNeXt-T. 
This demonstrates that our SwiftFormer models, powered by our proposed efficient additive attention, run faster than the lightweight CNN models on mobile devices and achieve superior performance. 
Recent device-level optimizations for CNN-based models, such as dedicated hardware implementations for convolutions with batch normalization and non-linearity, likely contribute to the high throughput of fully CNN-based models on A100.

\noindent
\textbf{Comparison with transformer models:}
Although transformer models usually outperform CNN-based models in terms of accuracy, they tend to suffer from high latency when running on resource-constrained mobile devices. For instance, DeiT-S, which has a similar model size to ResNet-50 and achieves higher top-1 accuracy by 3.3\%, but ResNet-50 runs approximately 5.2$\times$ faster on an iPhone 14 mobile device. In contrast, our SwiftFormer-L3 model achieves 1.2\% higher accuracy than DeiT-S, while running at the same speed as ResNet-50. Further our SwiftFormer-S model runs approximately 4.7$\times$ faster than DeiT-T on an iPhone 14 mobile device and has 6.3\% better accuracy.

\noindent
\textbf{Comparison with hybrid models:}
Although most existing hybrid approaches achieve higher accuracy compared to their lightweight CNN counterparts, they still under-perform the fully CNN-based models in terms of latency due to the quadratic complexity of multi-head self-attention. For example, EdgeViT-XXS runs at approximately 2$\times$ slower compared to MobileNet-v3-Large$\times$0.75. On the other hand, our SwiftFormer-XS has better latency as compared to lightweight CNNs and approximately is 2$\times$ faster than EdgeViT-XXS and MobileViT-XS, with an overall 1.3\% and 0.9\% higher top-1 accuracy respectively. Further, our SwiftFormer-L1 model is 3$\times$ faster than the state-of-the-art MobileViT-v2$\times$1.5 with 0.5\% better top-1 accuracy.~Our SwiftFormer-L3 model achieves 83.0\% top-1 accuracy and runs at 1.9 ms, which is 2.6$\times$ faster than MobileViT-v2$\times$2.0 with an absolute 1.8\% accuracy gain.

\begin{table*}[]
\centering
\small
\begin{tabular}{c|ccc|ccc||c}
\hline
\multirow{2}{*}{Backbone} & \multicolumn{6}{c||}{Detection \& Instance Segmentation }   &  Semantic \\
                          & $\textbf{AP}^{box}$  & $\textbf{AP}^{box}_{50}$ & $\textbf{AP}^{box}_{75}$ & $\textbf{AP}^{mask}$  & $\textbf{AP}^{mask}_{50}$ & $\textbf{AP}^{mask}_{75}$ &  mIoU($\%$)                \\
\hline
\hline
ResNet18~\cite{resnet}                  & 34.0 & 54.0  & 36.7  & 31.2 & 51.0  & 32.7  & 32.9                  \\
PoolFormer-S12~\cite{PoolFormer}            & 37.3 & 59.0  & 40.1  & 34.6 & 55.8  & 36.9  & 37.2                  \\
EfficientFormer-L1~\cite{EfficientFormer}           & 37.9 & 60.3  & 41.0 & 35.4 & 57.3 & 37.3 &    38.9               \\
\rowcolor{gray!10}\textbf{SwiftFormer-L1}           & \textbf{41.2} & \textbf{63.2}  & \textbf{44.8} & \textbf{38.1} & \textbf{60.2} & \textbf{40.7} &    \textbf{41.4}                   \\

\hline

ResNet50~\cite{resnet}                 & 38.0 & 58.6  & 41.4  & 34.4 & 55.1  & 36.7  & 36.7                  \\
PoolFormer-S24~\cite{PoolFormer}            & 40.1 & 62.2  & 43.4  & 37.0 & 59.1  & 39.6  & 40.3                  \\
EfficientFormer-L3~\cite{EfficientFormer}          &  41.4 & 63.9  &  44.7  &  38.1 &  61.0  & 40.4 & 43.5                  \\

\rowcolor{gray!10}\textbf{SwiftFormer-L3}           & \textbf{42.7} & \textbf{64.4}  & \textbf{46.7} & \textbf{39.1} & \textbf{61.7} & \textbf{41.8} &    \textbf{43.9}                   \\
\hline
\end{tabular}
\vspace{0.1cm}
\caption{
\textbf{Results using SwiftFormer as a backbone on dense prediction tasks}: Object detection and instance segmentation on COCO, whereas semantic segmentation on ADE20K. Our approach outperforms the recent EfficientFormer on all three tasks.}
\label{tab:detect_segmentation}
\end{table*}

\begin{figure*}[!h]
  \centering
    \includegraphics[width=1.0\linewidth]{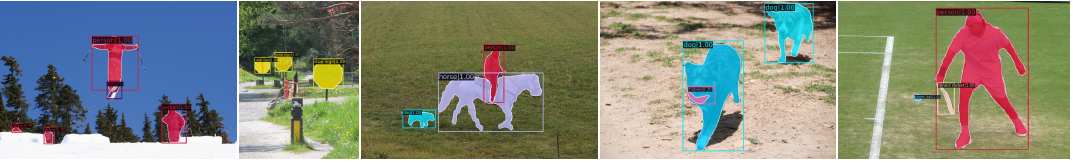}
    \caption{\textbf{Qualitative results on COCO.} The qualitative examples for object detection and instance segmentation on the COCO 2017 validation set. The visualizations show that our SwiftFormer-L1 based model can accurately detect and segment the instances in images.}
    \label{fig:detection_segmentation}
\end{figure*}

\begin{figure*}[!h]
  \centering
    \includegraphics[width=1.0\linewidth]{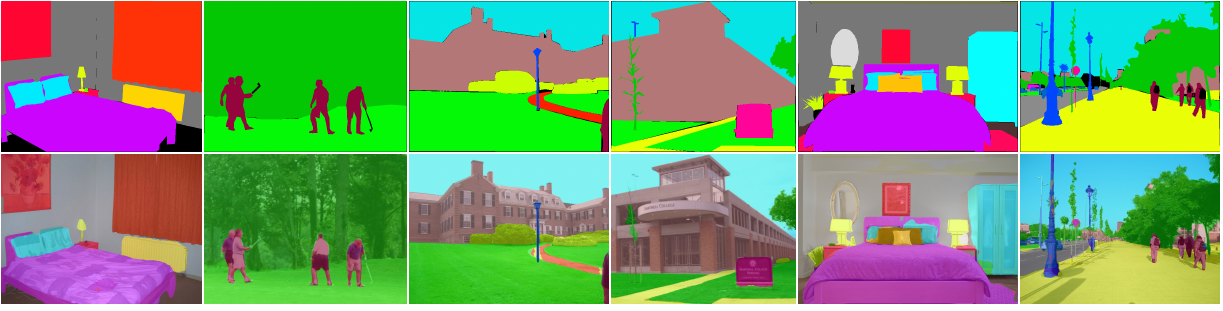}
    \caption{\textbf{Qualitative results on ADE20K.} The qualitative examples for semantic segmentation on the ADE20K validation set. \textbf{Top:} Ground truth masks. \textbf{Bottom:} The semantic segmentation results. Our model can accurately segment various indoor and outdoor scenes.}
    \label{fig:detection_segmentation}
\end{figure*}

\begin{figure}[!h]
  \centering
    \includegraphics[width=1.0\linewidth]{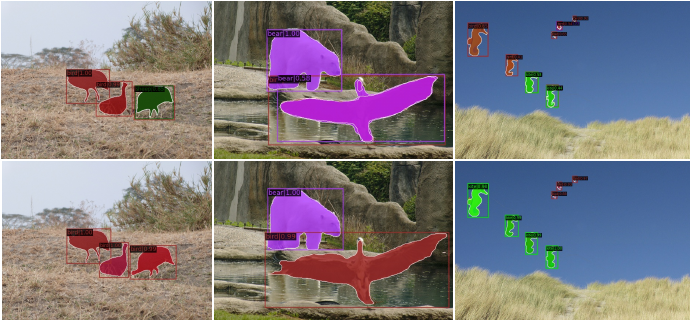}
    \caption{\textbf{Qualitative comparison with baseline.} Qualitative comparison between the recent EfficientFormer-L1 (\textbf{top}) and our SwiftFormer-L1 (\textbf{bottom}) on example images from COCO validation set for dense prediction tasks: detection and instance segmentation. Here, EfficentFormer-L1 misclassifies a bird as a sheep in the first example and as a bear in the second example, while misclassifying kites as birds in the third example. Our SwiftFormer-L1 accurately detects and segments objects in these examples.}
    \label{fig:detection_segmentation}
    \vspace{-0.5cm}
\end{figure}

\subsection{Object Detection and Instance Segmentation}
\noindent Table~\ref{tab:detect_segmentation}~compares the object detection and instance segmentation results of Mask-RCNN with different lightweight backbones.~Our SwiftFormer-L1 backbone achieves 41.2 AP box, surpassing the lightweight ResNet18 
and PoolFormer-S12 backbones by 7.2 and 3.9 points respectively. Further, it performs better than the previous state-of-the-art EfficientFormer-L1 backbone by 3.3 AP box. For instance segmentation, our method achieves 38.1 AP mask score which is 2.7 points better than the previous state-of-the-art. Similar trend is observed for SwinftFormer-L3 backbone, which surpasses the previous state-of-the-art EfficientFormer-L3 backbone by 1.3 points and 1.0 points in AP box and mask respectively. The improvement in the downstream detection and instance segmentation tasks illustrates the effectiveness of our SwiftFormer backbone models for the dense prediction tasks. 

\subsection{Semantic Segmentation}
\noindent Table~\ref{tab:detect_segmentation} shows the semantic segmentation results of SwiftFormer backbone-based models as compared to previously proposed backbones. We achieve 41.4\% mean intersection over union score using SwiftFormer-L1, surpassing ResNet18 by 8.5\%, PoolFormer-S12 by 4.2\%, and the state-of-the-art EfficientFormer-L1 by 2.5\%. Similarly, our SwiftFormer-L3 backbone-based segmentation model 
achieves 43.9 mIoU, surpassing all previous methods.

\section{Conclusion}
\noindent Transformers have gained popularity in vision applications due to their effective use of self-attention computation. However, their use in mobile vision applications is challenging due to the quadratic nature of the self-attention, which is computationally expensive on mobile devices.
To address this issue, many hybrid approaches and efficient variants of self-attention have been proposed.
In this work, we propose a novel efficient additive attention that replaces the expensive matrix multiplication operations with element-wise multiplications, and eliminates explicit keys-values interaction. Our proposed attention is linear with respect to the input tokens and can be used in all stages of the network. We show state-of-the-art results on image classification, object detection, and segmentation benchmarks.

{\small
\bibliographystyle{ieee_fullname}
\bibliography{egbib}
}

\clearpage
\newpage
\appendix

\begin{center}
\textbf{\Large Supplemental Material}
\end{center}

In this section, we provide additional details regarding:
\begin{itemize}\setlength{\itemsep}{-0.5em}
    \item  Architecture Details of SwiftFormer (Appendix~\ref{arch_impl})
    \item  Implementation Details (Appendix~\ref{impl_details})
    \item  Ablations (Appendix~\ref{additional_ablations})
    \item  Error Analysis on COCO Dataset (Appendix~\ref{error_analysis})
    \item  Qualitative Results (Appendix~\ref{qual_details})
    \item  Discussion (Appendix~\ref{discussion})
\end{itemize}


 \begin{table*}[hbt!]
 \centering
\scalebox{1}{
\small
\setlength{\tabcolsep}{7.5pt}
\renewcommand\arraystretch{1.0}

\begin{tabular}{c|c|c|c|c|c|c|c}
\toprule
\multirow{2}{*}{Stage} & \multirow{2}{*}{Output Resolution} & \multirow{2}{*}{Type}                                                      & \multirow{2}{*}{Config} & \multicolumn{4}{c}{SwiftFormer}  \\ \cline{5-8}
                       &                             &                                                                            &                         & XS         & S        & L1 & L3       \\
\hline
\hline
\multirow{4}{*}{stem}  & \multirow{2}{*}{$\frac{H}{2}\times \frac{W}{2}$}        & \multirow{2}{*}{\begin{tabular}[c]{@{}c@{}}Patch\\ Embed.\end{tabular}} & Patch Size              & \multicolumn{4}{c}{$k=3\times 3, s=2$}      \\ \cline{4-8}
                       &                             &                                                                            & Embed. Dim.             & 24         & 24        & 24 & 32       \\ 
                       \cline{2-8}
                       & \multirow{2}{*}{$\frac{H}{4}\times \frac{W}{4}$}        & \multirow{2}{*}{\begin{tabular}[c]{@{}c@{}}Patch\\ Embed.\end{tabular}} & Patch Size,             & \multicolumn{3}{c}{$k=3\times 3, s=2$}      \\ \cline{4-8}
                       &                             &                                                                            & Embed. Dim.             & 48         & 48        & 48   & 64    \\
                       \hline 
\multirow{2}{*}{1}     & \multirow{2}{*}{$\frac{H}{4}\times \frac{W}{4}$}        & \multirow{2}{*}{$\texttt{Hybrid}$}                                                      & Conv. Encoder [$C$, $N$]             & \multirow{1}{*}{48, 2} & \multirow{1}{*}{48, 2} & \multirow{1}{*}{48, 3} & \multirow{1}{*}{64, 3}  \\ 
                        &                             &                                                                            & SwiftFormer Encoder [$C$, $N$]               & \multirow{1}{*}{48, 1} & \multirow{1}{*}{48, 1} & \multirow{1}{*}{48, 1} & \multirow{1}{*}{64, 1}  \\ 
                       \hline 

\multirow{6}{*}{2}     & \multirow{2}{*}{$\frac{H}{8}\times \frac{W}{8}$}        & 

\multirow{2}{*}{\begin{tabular}[c]{@{}c@{}}Down-sampling\end{tabular}} & Patch Size              & \multicolumn{4}{c}{$k=3\times 3, s=2$}      \\ \cline{4-8}
                       &                             &                                                                            & Embed. Dim.             & 56         & 64       & 96   & 128   \\
                       \cline{1-8}
                       &  \multirow{2}{*}{$\frac{H}{8}\times \frac{W}{8}$}                           & \multirow{2}{*}{$\texttt{Hybrid}$}                                                      & Conv. Encoder [$C$, $N$]           & \multirow{1}{*}{56, 2} & \multirow{1}{*}{64, 2} & \multirow{1}{*}{96, 2} & \multirow{1}{*}{128, 3}  \\ 
                        &                             &                                                                            & SwiftFormer Encoder [$C$, $N$]                & \multirow{1}{*}{56, 1} & \multirow{1}{*}{64, 1} & \multirow{1}{*}{96, 1} & \multirow{1}{*}{128, 1} \\
                       \hline 
\multirow{6}{*}{3}     & \multirow{2}{*}{$\frac{H}{16}\times \frac{W}{16}$}       & \multirow{2}{*}{\begin{tabular}[c]{@{}c@{}}Down-sampling\end{tabular}} & Patch Size              & \multicolumn{4}{c}{$k=3\times 3, s=2$}      \\ \cline{4-8}
                       &                             &                                                                            & Embed. Dim.             & 112        & 168       & 192   & 320   \\
                       \cline{1-8}
                       &   \multirow{2}{*}{$\frac{H}{16}\times \frac{W}{16}$}                           & \multirow{2}{*}{$\texttt{Hybrid}$}                                                      & Conv. Encoder [$C$, $N$]           & \multirow{1}{*}{112, 5} & \multirow{1}{*}{168, 8} & \multirow{1}{*}{192, 9} & \multirow{1}{*}{320, 11}  \\ 
                        &                             &                                                                            & SwiftFormer Encoder [$C$, $N$]                    & \multirow{1}{*}{112, 1} & \multirow{1}{*}{168, 1} & \multirow{1}{*}{192, 1} & \multirow{1}{*}{320, 1}  \\
                       \hline 
\multirow{6}{*}{4}     & \multirow{2}{*}{$\frac{H}{32}\times \frac{W}{32}$}       & \multirow{2}{*}{\begin{tabular}[c]{@{}c@{}}Down-sampling\end{tabular}} & Patch Size              & \multicolumn{4}{c}{$k=3\times 3, s=2$}      \\ \cline{4-8}
                       &                             &                                                                            & Embed. Dim.             & 220        & 224       & 384   & 512   \\
                       \cline{1-8}
                       & \multirow{2}{*}{$\frac{H}{32}\times \frac{W}{32}$}                            & \multirow{2}{*}{$\texttt{Hybrid}$}                                                       & Conv. Encoder [$C$, $N$]              & \multirow{1}{*}{220, 3} & \multirow{1}{*}{224, 5} & \multirow{1}{*}{384, 4} & \multirow{1}{*}{512, 5}  \\ 
                        &                             &                                                                            & SwiftFormer Encoder [$C$, $N$]                & \multirow{1}{*}{220, 1} & \multirow{1}{*}{224, 1} & \multirow{1}{*}{384, 1} & \multirow{1}{*}{512, 1}  \\

\bottomrule
GMACs  &  &   &  & 0.6G & 1.0G & 1.6G & 4.0G\\
Parameters  &  &   &  & 3.5M & 6.1M & 12.1M & 28.5M\\
\bottomrule
\end{tabular}
}
\vspace{0.1cm}
\caption{\textbf{SwiftFormer Architectures.}~Description of the configurations of the model variants with respect to the output resolution, the output channels $C$, the number of blocks $N$, and the model's GMACs and parameters. Between two consecutive stages, we incorporate a downsampling layer to increase the number of channels and reduce the resolution by two.}
\label{tab:achitecture}
\end{table*}

\section{Architecture Details of SwiftFormer}
\label{arch_impl} 
\noindent
The detailed network architectures for SwiftFormer-XS, SwiftFormer-S, SwiftFormer-L1, and SwiftFormer-L3 are provided in Table~\ref{tab:achitecture}. We report the resolution, the number of channels ($C$) and the number of repeated blocks ($N$) of each stage for all the model variants. For all variants, we use an expansion ratio of 4 in the Conv.~Encoder. Since our architectures are not built using any neural architecture search, the number of channels and blocks for our models are selected to have a similar model size and GMACs with previous state-of-the-art methods in each variant.

\section{Additional Implementation Details}
\label{impl_details}
\noindent
We train and report the accuracy of our SwiftFormer models at 224${\times}$224 resolution for a fair comparison with the baseline and previous methods. We use a batch size of 2048 during training. The experiments for the SwiftFormer models were conducted on eight A100 GPUs, with an average training time of 36 hours for the classification. To enhance the robustness of the models, we apply several data augmentations during training. Specifically, we employed color jitter with a ratio of 0.4, RandAugment~\cite{RandAugment} with a magnitude of 9 and standard deviation of 0.5, gradient clipping of 0.01, Mixup~\cite{Mixup} and Cutmix~\cite{CutMix} with percentages of 1 and 0.8, respectively, label smoothing with a value of 0.1, and random erase with a probability of 0.25.
Similar to DeiT~\cite{deit} and EfficientFormer~\cite{EfficientFormer}, we employ RegNetY-16GF~\cite{RegNet} with 82.9\% top-1 accuracy as our teacher model for hard distillation.

\section{Additional Ablations}
\label{additional_ablations}
\noindent
We investigate the effect of QKV interactions and observe that eliminating key-value interactions and replacing them with a simple linear transformation results in 10\% reduction in latency. In addition to latency reduction, the top-1 accuracy is improved by 0.4\%. Overall, our results demonstrate the effectiveness of the proposed SwiftFormer encoder and highlight the potential benefits of simplifying the QKV interactions in the efficient additive attention mechanism.

\section{Error Analysis on COCO Dataset}
\label{error_analysis}

\begin{figure}[!h]
  \centering
    \includegraphics[width=0.95\linewidth]{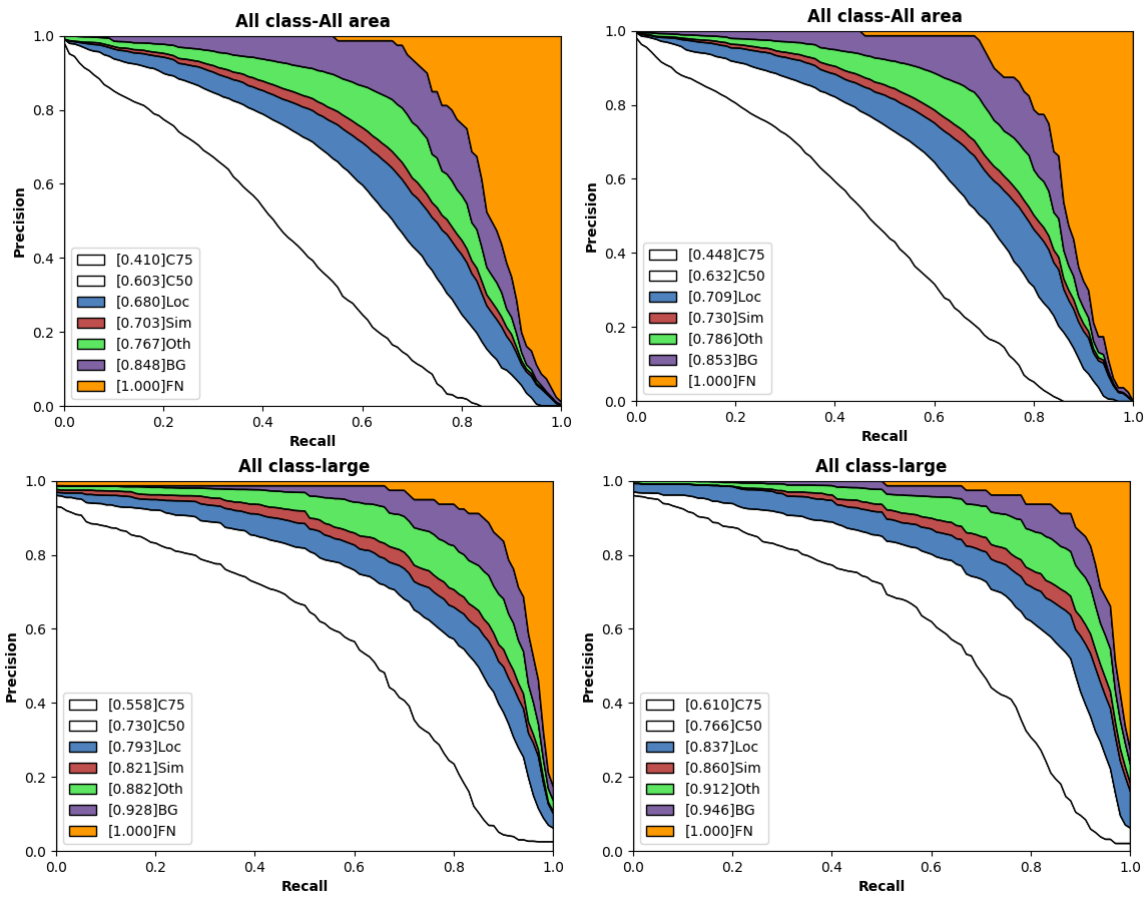}
    \caption{\textbf{Error analysis for the performance on COCO}. The baseline EfficientFormer-L1 (\textbf{left}) and our SwiftFormer-L1 (\textbf{right}) across all categories, on the all-objects (\textbf{top}) and large-sized objects (\textbf{bottom}). The plot in each image indicates a series of precision-recall curves using different evaluation configurations~\cite{COCO}, with the legend indicating the area under each curve in brackets. Our SwiftFormer-L1 provides consistent improvements over the baseline EfficientFormer-L1.}
    \label{fig:detection_analysis}
\end{figure}

\begin{figure*}[!t]
  \centering
    \includegraphics[width=0.95\linewidth]{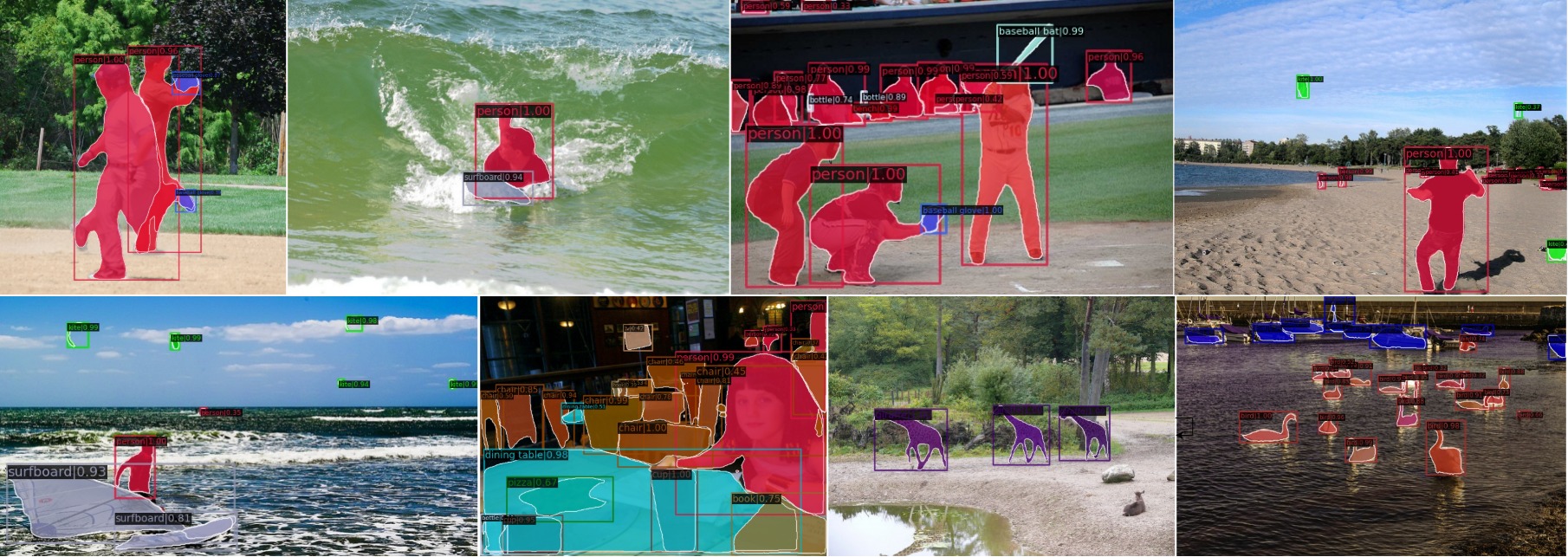}
    \caption{\textbf{Additional qualitative results on COCO.} Detection and instance segmentation results of our model.}
    \label{fig:sup_detection_segmentation}
\end{figure*}

Fig.~\ref{fig:detection_analysis} shows the error analysis plot of the baseline EfficientFormer-L1 (left) and our SwiftFormer-L1 (right) for all-objects and the large-sized objects.~We show the area under each curve in brackets in the legend. It is noted that our results are better compared to the baseline, especially for large-sized objects. For instance, the overall AP of large-sized objects of EfficientFormer-L1 at IoU=0.75 is 0.558 and perfect localization increases the AP to 0.793. Excluding the background false positives likely increase the performance to 0.928 AP. In the case of SwiftFormer, the overall AP at IoU=0.75 is 0.610 and perfect localization increases the AP to 0.837. Further, excluding the background false positives likely increase the performance to 0.946 AP. When analyzing the performance on small and medium-sized objects, we still have an improvement achieved by our SwiftFormer model.

\begin{figure*}[!t]
  \centering
    \includegraphics[width=1.0\linewidth]{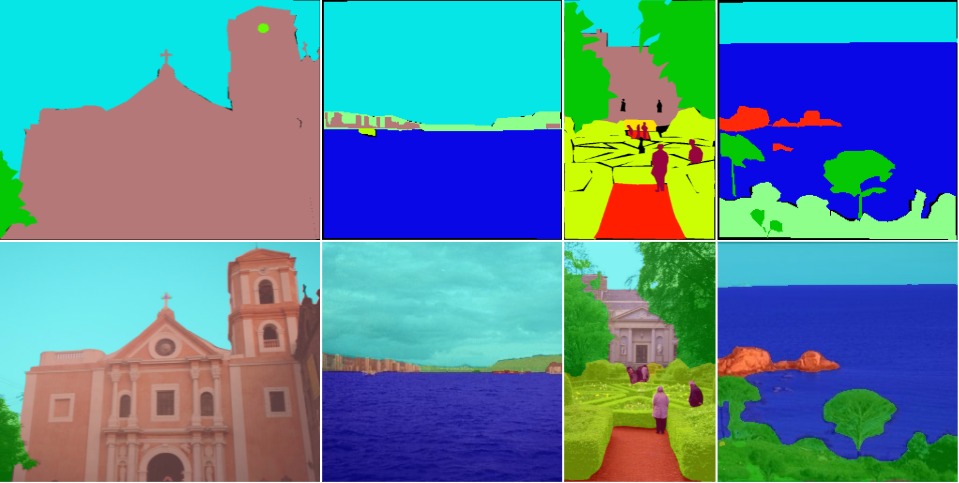}
    \caption{\textbf{Additional qualitative results on ADE20K.} Top row shows the ground truth masks and the bottom row shows the predictions of our model.}
    \label{fig:sup_semantic_segmentation}
\end{figure*}

\section{Qualitative Results}
\label{qual_details}
\noindent
Fig.~\ref{fig:sup_detection_segmentation} and~\ref{fig:sup_semantic_segmentation}  shows additional qualitative results of our SwiftFormer model for instance segmentation/detection and semantic segmentation respectively. Our model accurately localizes and segments the objects in diverse scenes. It also provides high-quality segmentation masks on ADE20K validation dataset.

\section{Discussion}
\label{discussion}
\noindent
The positional encoding and attention biases in vision transformers both play a crucial role in providing spatial information about the input sequence, particularly in dense prediction tasks. However, the attention bias is sensitive to input resolution and can make the model fragile when incorporated into these tasks. Meanwhile, typical positional encoding can slow down the inference of the model on resource-constrained devices. To overcome these challenges, we introduce an efficient additive attention mechanism that does not include positional encoding or attention biases, allowing for fast inference speed. In addition, our SwiftFormer models have shown promising results in downstream tasks. To the best of our knowledge, SwiftFormer is currently the most efficient hybrid architecture for real-time mobile vision applications.

\end{document}